\documentclass{article}
\usepackage{ijcai20}

\usepackage{times}
\usepackage{soul}
\usepackage{url}
\usepackage[hidelinks]{hyperref}
\usepackage[utf8]{inputenc}
\usepackage[small]{caption}
\usepackage{graphicx}
\usepackage{amsmath}
\usepackage{amsthm}
\usepackage{booktabs}
\usepackage{algorithm}
\usepackage{algorithmic}
\usepackage{bm}
\usepackage{amsfonts} 
\usepackage{mathabx}
\usepackage{subcaption}
\usepackage{color}
\usepackage{multirow}
\usepackage{stackengine}
\title{Compressing Recurrent Neural Networks Using Hierarchical Tucker Tensor Decomposition}

\author{Miao Yin$^1$\and Siyu Liao$^1$\and Xiao-Yang Liu$^2$ \and Xiaodong Wang$^2$ \and Bo Yuan$^1$ \affiliations $^1$Department of Electrical and Computer Engineering, Rutgers University\\ $^2$Department of Electrical Engineering, Columbia University \emails \{miao.yin, siyu.liao\}@rutgers.edu, \{xl2427, xw2008\}@columbia.edu, bo.yuan@soe.rutgers.edu}

\begin{document}

\maketitle

\begin{abstract}

Recurrent Neural Networks (RNNs) have been widely used in sequence analysis and modeling. However, when processing high-dimensional data,  RNNs typically require very large model sizes, thereby bringing a series of deployment challenges. Although the state-of-the-art tensor decomposition approaches can provide good model compression performance, these existing methods are still suffering some inherent limitations, such as restricted representation capability and insufficient model complexity reduction. To overcome these limitations, in this paper we propose to develop compact RNN models using Hierarchical Tucker (HT) decomposition. HT decomposition brings strong hierarchical structure to the decomposed RNN models, which is very useful and important for enhancing the representation capability. Meanwhile, HT decomposition provides higher storage and computational cost reduction than the existing tensor decomposition approaches for RNN compression. Our experimental results  show that, compared with the state-of-the-art compressed RNN models, such as TT-LSTM, TR-LSTM and BT-LSTM, our proposed HT-based LSTM (HT-LSTM), consistently achieves simultaneous and significant increases in both compression ratio and test accuracy on different datasets.

\end{abstract}

\section{Introduction}
\label{sec:intro}

Recurrent Neural Networks (RNNs), especially their advanced variants such as Long-Short Term Memory (LSTM) and Gated Recurrent Unit (GRU), have achieved unprecedented success in sequence analysis and processing. Thanks to their powerful capability of capturing and modeling the temporary dependency and correlation in the sequential data, the state-of-the-art RNNs have been widely deployed in many important artificial intelligence (AI) fields, such as natural language processing (NLP) \cite{sutskever2011generating}, speech recognition \cite{mikolov2011extensions}, and computer vision \cite{yu2016video}.

Despite their current prosperity, the efficient deployment of RNNs is still facing several challenges, especially the \textit{large model size problem}. Due to the widespread existence of high-dimensional input data in many applications, e.g. NLP and video processing, the input-to-hidden weight matrices of RNNs are often extremely large. For instance, as pointed out in \cite{yang2017tensor}, even with small-size hidden layer such as 256 hidden states, an LSTM working on UCF11 video recognition dataset \cite{liu2009recognizing} already requires more than 50 million parameters. Such ultra-high model size, consequently, brings a series of deployment challenges for RNNs, including but not limited to high difficulty of training, susceptibility to overfitting, long processing latency and inefficient energy consumption etc.

\textbf{Prior Work on RNN Compression.} To address RNNs' ultra-large model size problem, several \textit{model compression} approaches, such as pruning and  quantization \cite{han2015deep}, have been proposed and studied in prior work. Among them, the most promising solution is \textit{Tensor Decomposition}, a technique that represents a large tensor with the combination of multiple small tensor cores. By its nature, tensor decomposition approach is inherently a powerful tool for identifying and exploring the higher-order data correlation. From the perspective of model compression, such strong correlation-capturing capability makes tensor decomposition very attractive and well-suited for exploiting and reducing the model redundancy in large-scale RNNs. Recent advances in model compression research already show that, various tensor decomposition-based compression methods, including tensor train (TT) \cite{yang2017tensor}, tensor ring (TR) \cite{pan2019compressing} and block-term (BT) \cite{ye2018learning}, can bring several orders-of-magnitude fewer parameters for large-size RNNs with still maintaining high classification/prediction performance.

\textbf{Limitations of Prior Work.} Although the existing tensor decomposition-based RNN compression approaches already show their promising potentials, these state-of-the-art methods are still facing two inherent limitations: 1) the tensor decomposition approaches used in \cite{yang2017tensor}, \cite{ye2018learning} and \cite{pan2019compressing} have strict constraints on either the shapes or the combination manners of the component tensor cores, thereby limiting the representation ability of the corresponding compressed RNN models. For instance, TT decomposition requires the border tensor cores have to be rank-1, which directly hinders the representation power of TT-based RNNs. More generally, when using TT, TR or BT, the important hierarchical structure, which is important to capture many inherent hierarchical patterns or representation in the data, is missing at the inter-tensor-core level, thereby limiting the representation capability of the entire neural network models; and 2) from the perspective of complexity analysis, TT, TR and BT are not the best tensor decomposition approaches that provide the most promising space or computational complexity reduction. For instance, executing BT-based RNN models suffers computation overhead due to the extra flatten and permutation operations. More generally, when we consider to apply tensor decomposition to compress RNN models, in many settings the state-of-the-art TT, TR and BT solutions are inferior to some other types of tensor decomposition option in terms of numbers of parameters and/or operations saving. Consequently, the existing tensor decomposition-based compression methods are not the optimal solutions to fully exploit and minimize the RNN model redundancy.

\textbf{Technical Preview \& Benefits.} To overcome these limitations and fully unlock the potentials of tensor decomposition in model compression, in this work we propose to develop compact RNN models by using \textit{Hierarchical Tucker} (HT) decomposition \cite{hackbusch2009new}, a little explored but powerful tool for capturing and modeling the correlation and structure in the high-dimensional data. Unlike other popularly used tensor decomposition methods such as TT, TR and BT, HT decomposition enables the decomposed tensors exhibit strong hierarchical structure, which is very useful and important for enhancing the representation capability of RNN models. Meanwhile, comparing to its well-explored counterparts, HT decomposition can bring more complexity reduction with the same rank setting, thereby enabling HT-based RNN models have lower storage and computational costs than the prior work with the same or even higher accuracy. In overall, the features and benefits of HT-based RNN models are summarized as follows:

\begin{itemize}
   
    \item HT-based RNNs exhibit stronger representation power than the existing tensor decomposition-based models. More specifically, the hierarchical structure imposed on the input-to-hidden layers makes RNNs can exploit and extract the important representation and pattern from high-dimensional data in a much more hierarchical, precise and comprehensive way, thereby significantly improving RNN models' representation capability. This benefit on representation capability is also verified by the empirical experiments. Our proposed HT-LSTM, as the the compressed LSTM models using HT decomposition, achieves higher accuracy than vanilla RNN, the state-of-the-art compressed RNNs as well as non-RNN models on various video recognition datasets.
    
    \item HT-based RNN models have much lower storage and computational costs than the state of the art. Compared with TT, TR and BT decomposition adopted in prior work, HT decomposition inherently provides higher complexity reduction on the same-size tensor data with the same selected rank. By leveraging such theoretical advantage, we can compress large-size RNN models in the HT format with requiring very few parameters. Experimental results show, compared with vanilla LSTM, HT-LSTM achieves very high compression ratio with even higher accuracy. Meanwhile, compared with the state-of-the-art compressed LSTM such as TT-LSTM, TR-LSTM and BT-LSTM, HT-LSTM consistently achieve simultaneous and significant increase in both compression ratio and test accuracy on different datasets.

\end{itemize}


\section{Hierarchical Tucker-based RNN Models}
\label{sec:HTRNN}

In this section, we describe the details of HT-based RNN models. First, we introduce the preliminaries of tensor basics, tensor computation and hierarchical tucker decomposition. Then, we reformulate the forward propagation and backward propagation procedure on the original input-to-hidden layer to the HT format, and thereby forming a new HT layer, which is the building component of our proposed compact HT-based RNN models. Furthermore, in order to evaluate the efficiency of HT-based compression approach, we analyze the computational and storage complexity of HT-based RNN models and make comparison with the other methods.

\subsection{Preliminaries}
\label{subsec:prelim}

\begin{figure}[t]
    \centering
    \includegraphics[width=\linewidth]{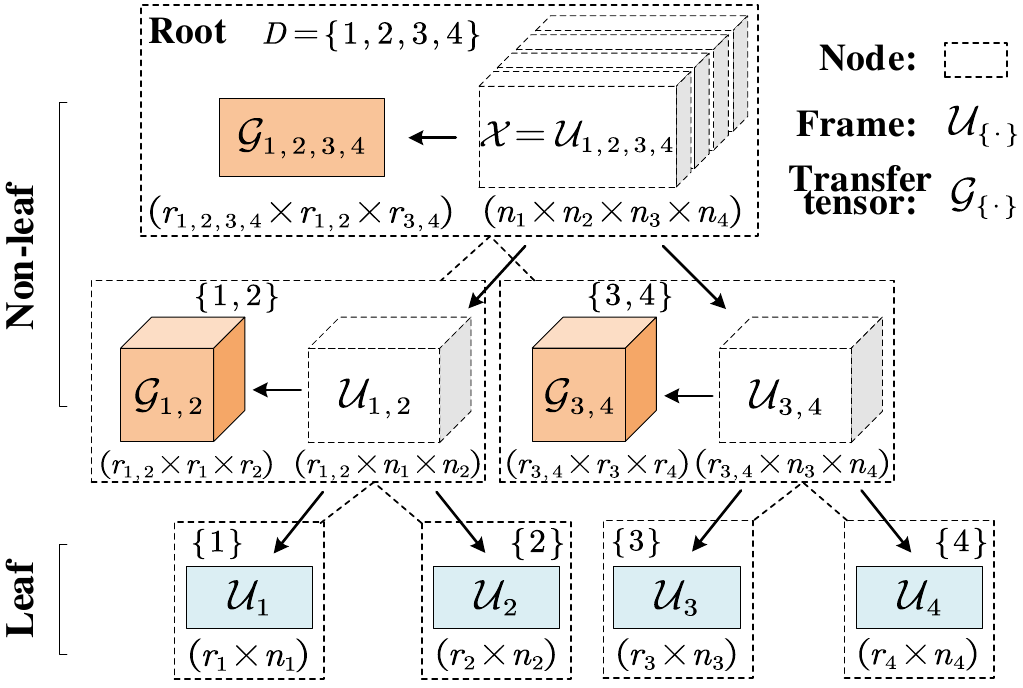}
    \caption{An illustration of HT decomposition with $d$=4. All the dashed lines and boxes describe a binary tree with root $D=\{1, 2, 3, 4\}$, where the dashed boxes represent the nodes. Here node $\{1\}$ is a leaf node, whose father and brother are node $\{1,2\}$ and node $\{2\}$, respectively. For leaf nodes, they are only associated with leaf frame, and for non-leaf nodes, they are associated with transfer tensors and non-leaf frames. Here $\bm{\mathcal{X}}$ is decomposed to a set of orange-colored transfer tensors and blue-colored leaf frames. }
    \label{fig:htd_example}
    \vspace{-4.6mm}
\end{figure}

\begin{figure*}[ht!]
    \centering
    \includegraphics[width=\linewidth]{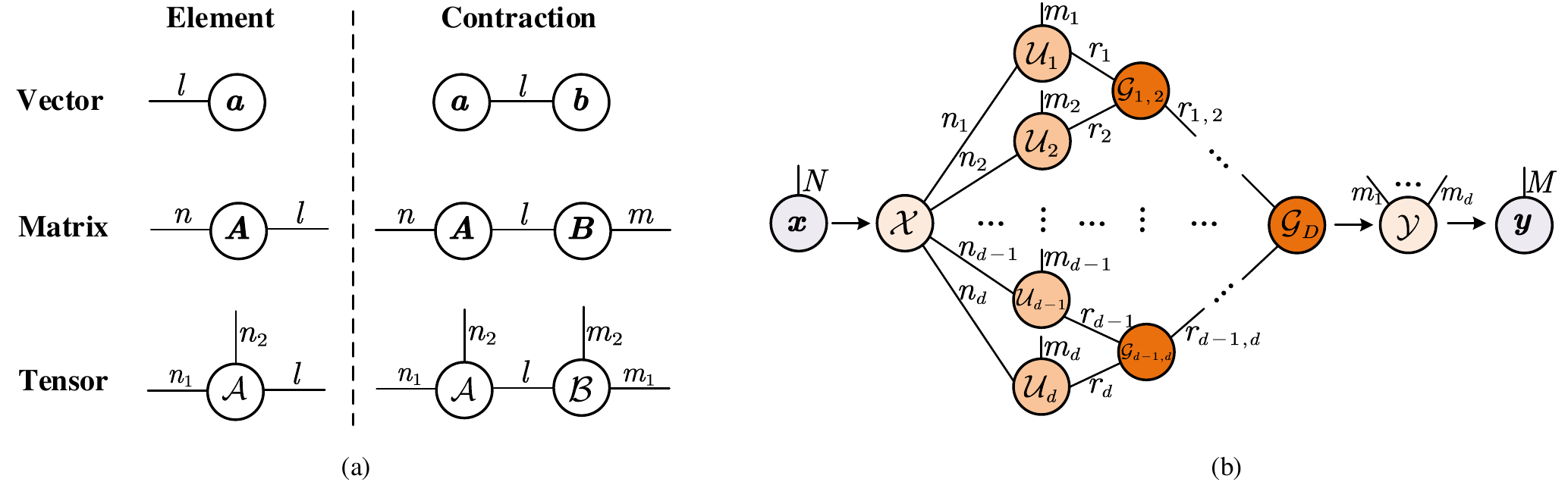}
    \vskip -2mm
    \caption{(a) Graphical representation of element and computation in the tensor diagram. Here the each line represents one dimension, and the variable near the line is the size of that dimension. (b) Representation of matrix-vector multiplication on a HT-decomposed layer using tensor diagram. Here weight matrix is already decomposed to the HT format with a set of 2-D leaf frames and 3-D transfer tensor.}
    \label{fig:htl}
    \vspace{-4.5mm}
\end{figure*}

\indent \textbf{Notation.} Throughout the paper we use boldface calligraphic script letters, boldface capital letters, and boldface lower-case letters to represent tensors, matrices, and vectors, respectively, e.g.  $\bm{\mathcal{X}}\in\mathbb{R}^{n_1 \times n_2 \times \cdots \times n_d}$,  $\bm{X}\in\mathbb{R}^{n_1\times n_2}$, and $\bm{x}\in\mathbb{R}^{n_1}$. In addition,  $\bm{\mathcal{X}}_{(i_1,\cdots,i_d)}\in\mathbb{R}$ denotes the entry of tensor  $\bm{\mathcal{X}}$. 
Similarly, $\bm{X}_{(i,j)}$ represents the entry of matrix $\bm{X}$.

 \textbf{Tensor Contraction.} An HT-decomposed tensor is essentially the consecutive product of multiple tensor contraction results, where tensor contraction is executed between two tensors with at least one matched dimension. For instance, given two tensors $\bm{\mathcal{A}}\in\mathbb{R}^{n_1\times n_2 \times l}$ and $\bm{\mathcal{B}}\in\mathbb{R}^{l\times m_1\times m_2}$, where the 3rd dimension of $\bm{\mathcal{A}}$ matches the 1st dimension of $\bm{\mathcal{B}}$ with length $l$, the tensor contraction result is a size- $n_1\times n_{2}\times m_1 \times m_2$ tensor as
$$
(\bm{\mathcal{A}}\times_{1}^{3}\bm{\mathcal{B}})_{(i_1, i_2, j_1, j_2)}=\sum_{\alpha=1}^{l}\bm{\mathcal{A}}_{(i_1, i_2, \alpha)}\bm{\mathcal{B}}_{(\alpha, j_1, j_2)}.
$$

\textbf{Hierarchical Tucker Decomposition.} The Hierarchical Tucker decomposition is a special type of tensor decomposition approach with hierarchical levels with respect to the order of the tensor. As illustrated in Figure \ref{fig:htd_example}, an HT-decomposed tensor can be recursively decomposed into intermediate components, referred as \textit{frames}, from top to bottom in a binary tree, where each frame corresponds to a unique \textit{node}, and each node is associated with a \textit{dimension set}. 
In general, for a HT-decomposed tensor $\bm{\mathcal{X}}\in\mathbb{R}^{n_1\times\cdots\times n_d}$, we can build a binary tree with a root node associated with $D=\{1,2,\cdots,d\}$ and $\bm{\mathcal{X}}=\bm{\mathcal{U}}_D$ as the root frame. 
For each non-leaf frame $\bm{\mathcal{U}}_{s}\in\mathbb{R}^{r_{s}\times n_{\mu_s}\times\cdots\times n_{\nu_s}}$, where $s\subsetneq D$ is associated with the node corresponding to $\bm{\mathcal{U}}_s$ , $s_1, s_2\subsetneq s$ are associated with the left and right child nodes of the $s$-associated node, and $\mu_s=\min(s), \nu_s=\max(s)$, $\bm{\mathcal{U}}_{s}$ can be recursively decomposed to its left and right child frames ($\bm{\mathcal{U}}_{s_1}$ and $\bm{\mathcal{U}}_{s_2}$) and transfer tensor $\bm{\mathcal{G}}_s\in\mathbb{R}^{r_s\times r_{s_1} \times r_{s_2}}$ as
\begin{equation}
\bm{\mathcal{U}}_s=\bm{\mathcal{G}}_s\times_{1}^{2}\bm{\mathcal{U}}_{s_1}\times_{1}^{2}\bm{\mathcal{U}}_{s_2}.
\label{eqn:htd}
\end{equation}

Consequently, by performing this recursive decomposition till the bottom of the binary tree, we can decompose the original ${n_1\times\cdots\times n_d}$-order tensor $\bm{\mathcal{X}}=\bm{\mathcal{U}}_D$ into the combination of the 2-order leaf frames and 3-order transfer tensors. Notice that here $r_s$, as \textit{hierarchical rank}, is an important parameter that determines the decomposition effect.

\subsection{HT-based RNN Models}
\label{subsec:HTRNN}

This subsection describes the details of compressing the large-size RNN models to the compact ones using HT decomposition, including the key steps as well as the important HT-based forward and backward propagation schemes.

\textbf{Tensorization.} In general, the key idea of building HT-RNN is to transform the weight matrix $\bm{W}\in\mathbb{R}^{M\times N}$ to the HT-based format. Considering $\bm{W}$ is a 2-D matrix, while HT decomposition is mainly performed on high-order tensor, we first need to reshape $\bm{W}$ as well as its affiliated input vector $\bm{x}\in\mathbb{R}^{N}$ and output vector $\bm{y}\in\mathbb{R}^{M}$ to tensor format as $\bm{\mathcal{W}}\in\mathbb{R}^{m_1\times\cdots\times m_d\times n_1\times\cdots\times n_d}$,  $\bm{\mathcal{X}}\in\mathbb{R}^{n_1\times\cdots\times n_d}$ and $\bm{\mathcal{Y}}\in\mathbb{R}^{m_1\times\cdots\times m_d}$, respectively, where $M=\prod_{i=1}^{d}m_i$ and $N=\prod_{j=1}^{d}n_j$.


\textbf{Decomposing  $\bm{W}$.} Given a tensorized $\bm{W}$ as $\bm{\mathcal{W}}\in\mathbb{R}^{m_1\times\cdots\times m_d\times n_1\times\cdots\times n_d}$, we can now leverage HT decomposition to represent the large-size $\bm{W}$ using a set of small-size matrices and tensors. In general, following Equation (\ref{eqn:htd}), $\bm{\mathcal{W}}$ can be decomposed as
\begin{equation}
\begin{aligned}
\bm{\mathcal{W}}_{(i_1,\cdots,i_d,j_1,\cdots,j_d)}=
\sum_{k=1}^{r_D}\sum_{p=1}^{r_{D_1}}\sum_{q=1}^{r_{D_2}}(\bm{\mathcal{G}}_D)_{(k,p,q)}\\
\cdot(\bm{\mathcal{U}}_{D_1})_{(p,\varphi_{D_1}(\bm{i},\bm{j}))}(\bm{\mathcal{U}}_{D_2})_{(q,\varphi_{D_2}(\bm{i},\bm{j}))},
\label{eqn:w_ht}
\end{aligned}
\end{equation}
where $\varphi_s(\bm{i},\bm{j})$ is a mapping function that produces the correct indices $\bm{i}=(i_1,\cdots,i_d)$ and $\bm{j}=(j_1,\cdots, j_d)$ for a specified frame $\bm{\mathcal{U}}_s$ with the given $s$ and $d$. For instance, with $d=6$ and $s=\{3,4\}$, the output of $\varphi_s(\bm{i},\bm{j})$ is $(i_3,i_4,j_3,j_4)$. In addition, $\bm{\mathcal{U}}_{D_1}$ and $\bm{\mathcal{U}}_{D_2}$ can be recursively computed as
\begin{equation}
\begin{aligned}
(\bm{\mathcal{U}}_s)_{(k,\varphi_s(\bm{i},\bm{j}))}
=\sum_{p=1}^{r_{s_1}}\sum_{q=1}^{r_{s_2}}(\bm{\mathcal{G}}_{s})_{(k,p,q)}\\
\cdot(\bm{\mathcal{U}}_{s_1})_{(p,\varphi_{s_1}(\bm{i},\bm{j}))}(\bm{\mathcal{U}}_{s_2})_{(q,\varphi_{s_1}(\bm{i},\bm{j}))},
\end{aligned}
\end{equation}
where $D=\{1,2,\cdots,d\}$, $D_1=\{1,\cdots,\lfloor d/2\rfloor\}$ and $D_2=\{\lceil d/2\rceil,\cdots,d\}$ are associated with left and right child nodes of the root node.


\textbf{HT Layer.} With the HT-decomposed weight matrix, the component HT layer of HT-based RNN models can now be developed. Specifically, the HT-format matrix-vector multiplication, as the kernel computation in the forward propagation procedure, is performed as follows:
\begin{equation}
\begin{aligned}
\bm{\mathcal{Y}}_{(\bm{i})}=\sum_{\bm{j}}
\sum_{k=1}^{r_D}\sum_{p=1}^{r_{D_1}}\sum_{q=1}^{r_{D_2}}(\bm{\mathcal{G}}_D)_{(k,p,q)}\\
\cdot(\bm{\mathcal{U}}_{D_1})_{(p,\varphi_{D_1}(\bm{i},\bm{j}))}(\bm{\mathcal{U}}_{D_2})_{(q,\varphi_{D_2}(\bm{i},\bm{j}))}\bm{\mathcal{X}}_{(\bm{j})}.
\end{aligned}
\end{equation}

Considering the desired output $\bm{y}$ of HT-layer is a vector, the calculated $\bm{\mathcal{Y}}$ needs to be re-shaped again to the 1-D format. Consequently, we denote the entire forward computing procedure from input $\bm{x}$ to output $\bm{y}$ as
\begin{equation}
\bm{y}= HTL(\bm{W}, \bm{x}).
\end{equation}

\textbf{Remark-1: Hierarchical Structure in HT-layer.} In tensor theory both the tensors and their computations can be graphically represented using \textit{tensor diagram} (see Figure \ref{fig:htl}(a)). For HT decomposition, its inherent hierarchical characteristics make the corresponding HT-layer exhibit strong multi-level hierarchical structure, which is visualized in the specific tensor diagram of HT-layer (see Figure \ref{fig:htl}(b)).
Considering that a well-known feature and advantage of deep neural network is its strong ability of capturing hierarchical pattern and representation via its multi-layer structure, the existence of such hierarchical structure in HT-layer can effectively improve the representation capability of the overall neural network models.
In Section \ref{sec:exp}, empirical experiments on different datasets demonstrate that HT-based RNN models indeed outperform many other types of RNN models in terms of accuracy. Besides, it is worth noting that some recent advances in learning theory \cite{nadav2018analysis} also indicates the strong connection and correlation between HT modeling and expressive power of deep neural networks.



\begin{table}
\setlength\tabcolsep{4.6pt}
\def\arraystretch{1.15}
\centering
\begin{tabular}{|c|l|l|} 
\hline

\textbf{Model}  & \multicolumn{1}{c|}{\textbf{Space}}  &  \multicolumn{1}{c|}{\textbf{Time}}   \\
\hline\hline
RNN FP &  \multirow{2}{*}{$\mathcal{O}(NM)$} & $\mathcal{O}(NM)$  \\
RNN BP &   & $\mathcal{O}(NM)$   \\
\hline
TT-RNN FP &  \multirow{2}{*}{$\mathcal{O}(dmnr^2)$} & $\mathcal{O}(dmr^2N)$  \\
TT-RNN BP &   & $\mathcal{O}(d^2mr^4N)$  \\
\hline
TR-RNN FP &  \multirow{2}{*}{$\mathcal{O}(dmnr^2)$} & $\mathcal{O}(dr^3N+dr^3M)$  \\
TR-RNN BP &   & $\mathcal{O}(d^2r^5N+nd^2r^5M)$   \\
\hline
BT-RNN FP &  \multirow{2}{*}{$\mathcal{O}(dmnr+r^d)$} & $\mathcal{O}(dmr^dNC)$   \\
BT-RNN BP &  & $\mathcal{O}(d^2mr^dNC)$   \\
\hline
HT-RNN FP & \multirow{2}{*}{$\mathcal{O}(dmnr+dr^3)$} & $\mathcal{O}(dmr^2N+dr^3N)$  \\
HT-RNN BP &  & $\mathcal{O}(d^2mr^5N+d^2r^6N)$   \\

\hline
\end{tabular}
\caption{Comparison of complexity with different tensor decomposition-based RNNs. Here FP and BP mean forward and backward propagation, respectively. $C$ is the CP rank-value defined in BT decomposition, and $r=\max_{s\subsetneq D}r_s$, $m=\max_{k\in D}m_{k}$, $n=\max_{k\in D}n_{k}$,}
\label{tab:complexity}
\vspace{-4.6mm}
\end{table}

\textbf{HT-LSTM.} With the HT-based component layers, a HT-based RNN model can be simply constructed by replacing the original uncompressed layer with the HT-layer. Considering LSTM is the most popular and advanced variant of RNNs, we develop the corresponding HT-LSTM model as follows:
\begin{equation}
\begin{aligned}
\bm{u}[t]=&\sigma(HTL(\bm{W}_u,\bm{x}[t])+\bm{V}_u\bm{h}[t-1]+\bm{b}_u)\\
\bm{f}[t]=&\sigma(HTL(\bm{W}_f,\bm{x}[t])+\bm{V}_f\bm{h}[t-1]+\bm{b}_f)\\
\bm{o}[t]=&\sigma(HTL(\bm{W}_o,\bm{x}[t])+\bm{V}_o\bm{h}[t-1]+\bm{b}_o)\\
\bm{c}[t]=&\bm{f}[t]\odot\bm{c}[t-1]+\bm{u}[t]\odot\\
&\tanh(HTL(\bm{W}_c,\bm{x}[t])+\bm{V}_c\bm{h}[t-1]+\bm{b}_c)\\
\bm{h}[t]=&\bm{o}[t]\odot\tanh(\bm{c}[t]),
\end{aligned}
\end{equation}
where $\sigma$, $\tanh$ and $\odot$ are the sigmoid function, hyperbolic function and element-wise product, respectively.

\textbf{HT-based Gradient Calculation.} To ensure the valid training on HT-RNN, the gradient calculation in the backward propagation should also be accordingly reformulated to HT-based format. In general, considering for HT layer $\bm{\mathcal{W}}=\bm{\mathcal{U}}_D$ and $\frac{\partial\bm{\mathcal{Y}}}{\partial\bm{\mathcal{U}}_D}=\bm{\mathcal{X}}$, assuming $s$ is associated with a left node, as we denote that $F(s)$ and $B(s)$ are the sets associated with the father and brother nodes of the $s$-associated node in the binary tree, respectively, and define  $\mu_s=\min(s), \nu_s=\max(s)$, the partial derivative of output tensor with respect to frames can be calculated in the following recursive way until $F(s)$ is equal to $D$:
\begin{equation}
\begin{aligned}
\frac{\partial\bm{\mathcal{Y}}}{\partial\bm{\mathcal{U}}_s}=&
\bm{\mathcal{G}}_{F(s)}\times_{1}^{3}\bm{\mathcal{U}}_{B(s)}\\
&\times_{\substack{1,\cdots,\nu_{B(s)}-\mu_{B(s)}+2,\nu_{F(s)}-\mu_{F(s)}+3,\\\cdots,\nu_{F(s)}-\mu_{F(s)}+\nu_{B(s)}-\mu_{B(s)}+3}}^{1,3,\cdots,2\nu_{B(s)}-2\mu_{B(s)}+4}\frac{\partial\bm{\mathcal{Y}}}{\partial\bm{\mathcal{U}}_{F(s)}}.
\end{aligned}
\label{eqn:rec_derivative}
\end{equation}
Based on Equation (\ref{eqn:rec_derivative}), the gradients for leaf frames and transfer tensors can be computed as follows: 
\begin{align}
\frac{\partial L}{\partial\bm{\mathcal{U}}_s}=&\frac{\partial\bm{\mathcal{Y}}}{\partial\bm{\mathcal{U}}_s}\times_{1,\cdots,\mu_s-1,\nu_s+1,\cdots,d}^{\nu_s-\mu_s+3,\cdots,d+1}\frac{\partial L}{\partial\bm{\mathcal{Y}}},\label{eqn:gradients_u}\\
\frac{\partial L}{\partial\bm{\mathcal{G}}_s}=&\frac{\bm{\mathcal{Y}}}{\partial\bm{\mathcal{U}}_s}\times_{2,\cdots,\nu_{s_1}-\mu_{s_1}+2}^{2,\cdots,\nu_{s_1}-\mu_{s_1}+2} \bm{\mathcal{U}}_{s_1}\nonumber\\
&\times_{2,\cdots,\nu_{s_2}-\mu_{s_2}+2}^{3,\cdots,\nu_{s_2}-\mu_{s_2}+3}\bm{\mathcal{U}}_{s_2}\times_{1,\cdots,d}^{4,\cdots,d+3}\frac{\partial L}{\partial \bm{\mathcal{Y}}}.\label{eqn:gradients_g}
\end{align}


\subsection{Complexity Analysis}
\label{subsec:complexity} 

To better understand the impact of HT decomposition on RNNs, we analyze the theoretical complexity of HT-RNN, and compare it with the vanilla uncompressed RNN as well as other tensor decomposition-based RNN models. Table \ref{tab:complexity} summarizes the space complexity and time complexity of different RNN models. It is seen that compared with the other compressed RNN models using tensor decomposition, HT-RNN has the lowest space complexity to store the model parameters. Meanwhile, HT-RNN also enjoys less time complexity than most listed models for both forward and backward propagation. It is worth noting that though TT-RNN has even less time complexity than HT-RNN, it has weaker representation capability, which translates to the lower accuracy, as demonstrated via the experimental results in Section \ref{sec:exp}.

\begin{figure}[t]
\vspace{-2ex}
    \centering
    \includegraphics[width=\linewidth]{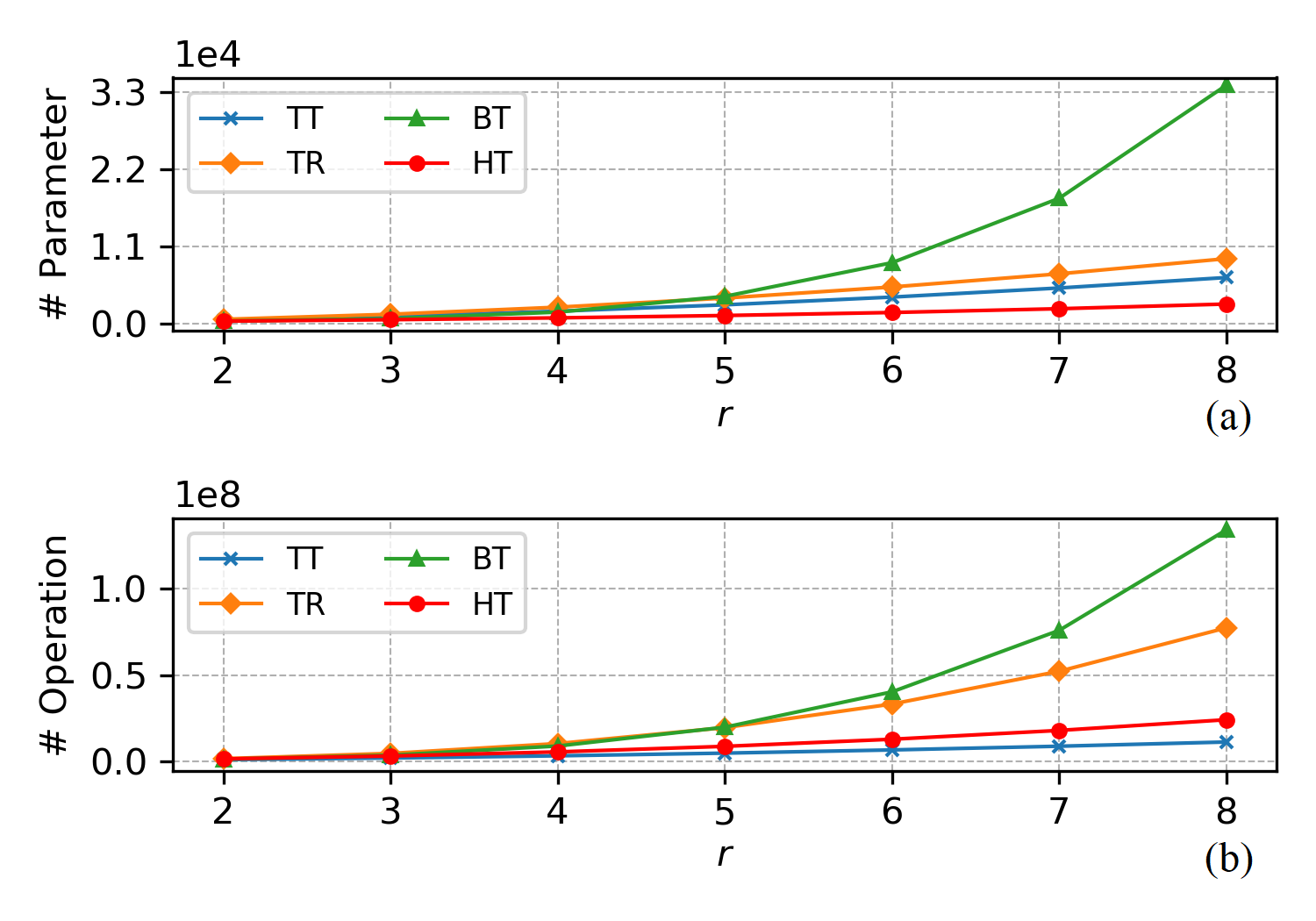}
    \vskip -3ex
    \caption{Top: Comparison on number of required parameters of weight matrix. Bottom: Comparison on number of required operations of matrix-vector multiplications. All the tensor decomposition methods use the same setting $d=5$, $(n_1,\cdots,n_5)=(8,10,10,9,8)$, $(m_1,\cdots,m_5)=(4,4,2,4,2)$, and $r$ is the rank.}
    \label{fig:complexity}
    \vspace{-4.6mm}
\end{figure}

\textbf{Remark-2.} Besides theoretical complexity analysis, we also verify the low-cost benefits of HT-based compression via empirical experiments. Figure \ref{fig:complexity}(a) shows the number of parameters to store a compressed weight matrix, and Figure \ref{fig:complexity}(b) shows the number of needed operations for multiplication between the compressed weight matrix and vector. Here we adopt the size-$57,600 \times256$ weight matrix used in \cite{yang2017tensor} \cite{ye2018learning} \cite{pan2019compressing} for evaluation. From Figure \ref{fig:complexity} it is seen that HT-based approach indeed achieves lower costs, especially on storage requirement, than other tensor decomposition-based methods. Next, the experiments in Section \ref{sec:exp} further show the advantages of HT-based RNN on compression ratios over various datasets.

\section{Experiments}
\label{sec:exp}

In this section, we evaluate the performance of HT-based RNN on different datasets, and compare them with the state-of-the-art with respect to compression ratio and test accuracy. Considering LSTM is the current most commonly used RNN variant in both academia and industry, our experiments focus on HT-LSTM, and compare it with the vanilla uncompressed LSTM and recent advances in compressed RNN such as TT-LSTM, BT-LSTM and TR-LSTM. Besides, we also compare HT-LSTM with other reported models that can be evaluated on the same datasets.



\textbf{Training Strategy.} Following the similar setting in prior work, we adopt two types of training strategy: end-to-end direct training and training with pre-trained CNN. In the end-to-end direct training the input of LSTM is the raw data, e.g. video clips; while training with pre-trained CNNs means the back-end LSTM receives the compact features extracted by a front-end pre-trained CNN. Next we describe our experiments belonging to these two categories, respectively.

\subsection{End-to-End Direct Training}

\textbf{Hyperparameter Setting.} We train the models using ADAM optimizer with L2 regularization of coefficient 0.001. Also, dropout rate is set as 0.25 and batch size is 16.

\textbf{UCF11 Dataset.} The UCF11 dataset \cite{liu2009recognizing} consists of 11-class human action (e.g. biking, diving, basketball) videos with totally 1,600 video clips. Each class is assembled by 25 video groups, where each group contains at least 4 action clips. For each clip, the resolution is $320\times 240$. 

At data pre-processing stage we choose the same settings used in the related work \cite{ye2018learning} \cite{pan2019compressing} for fair comparison. Specifically, the resolution of video clips is first scaled down to $160 \times 120$, and then 6 frames from each clip are randomly sampled to form the sequential input for our HT-LSTM model.

For the baseline vanilla uncompressed LSTM model, it contains 4 input-to-hidden layers, where the size of its input vector is $160\times 120\times 3=57,600$, and the number of hidden states in each layer is 256. For our proposed HT-LSTM, the input vector is reshaped to a tensor of shape $8\times 10\times 10\times 9\times 8$, and the output tensor of the input-to-hidden layer is of shape $4\times 4\times 2\times 4\times 2$. All leaf ranks are set as 4, and all non-leaf ranks are set as 5.

\begin{table}[h]
\centering
\def\arraystretch{1.15}
\begin{tabular}{|c|c|c|c|}
\hline
\textbf{Model}  & \textbf{CR} & \textbf{\# Param.} & \textbf{Accuracy} (\%) \\
\hline
\hline
LSTM  & 1     & 59M  & 69.7     \\
\hline
TT-LSTM  & \multirow{2}{*}{17,554$\times$} & \multirow{2}{*}{3,360}  & \multirow{2}{*}{79.6}    \\
(ICML-17) & & & \\
\hline
BT-LSTM  & \multirow{2}{*}{17,414$\times$}  & \multirow{2}{*}{3,387}  & \multirow{2}{*}{85.3}     \\
(CVPR-18) & & & \\
\hline
TR-LSTM  & \multirow{2}{*}{34,193$\times$}  & \multirow{2}{*}{1,725}  & \multirow{2}{*}{86.9}     \\
(AAAI-19) & & &\\
\hline
HT-LSTM   & \multirow{2}{*}{\textbf{47,375$\times$}} & \multirow{2}{*}{\textbf{1,245}}  & \multirow{2}{*}{\textbf{87.2}}     \\
(Ours) & & &\\
\hline
\end{tabular}
\caption{
Performance of different RNN compression work on UCF11 dataset using end-to-end direct training. CR stands for compression ratios. Results of TT-LSTM, BT-LSTM and TR-LSTM are reported in \protect\cite{yang2017tensor}, \protect\cite{ye2018learning} and  \protect\cite{pan2019compressing}, respectively.}
\label{tab:direct_ucf11}
\vspace{-5mm}
\end{table}

Table \ref{tab:direct_ucf11} summarizes the performance of our HT-LSTM on UCF11 dataset and compare it with the related work. It is seen that compared with vanilla LSTM using 59 million parameters, HT-LSTM only needs 47,375$\times$ fewer parameters with 17.5\% accuracy increase. Compared with the recent advances on compressing RNNs using other tensor decomposition methods, including TT-LSTM, BT-LSTM and TR-LSTM, our proposed HT-LSTM requires at least 1.38$\times$ fewer parameters with at least 0.3\% increase in accuracy.

\textbf{Youtube Celebrities Face Dataset.} Youtube dataset \cite{kim2008face} contains 1,910 video clips from 47 subjects, where each of it is a celebrated individual such as movie star. Also, the resolutions of the frames vary for different video clips. Being consistent with prior work, for data pre-processing the resolution of the input data to HT-LSTM is re-scaled as $160\times 120$. Also, 6 frames in each video clips are randomly sampled to form the input sequence. 

In this experiment we build a HT-LSTM with the similar setting with the one used in UCF11 dataset. To be specific, 4 input-to-hidden layers are equipped, and the shapes of tensorized input and output vectors are $8\times 10\times 10\times 9\times 8$ and $4\times 4\times 2\times 4\times 2$, respectively. A slight difference is in this HT-LSTM all leaf ranks are set as 3, and all non-leaf ranks are set as 4.

Table \ref{tab:directly_ytc} compares the performance of HT-LSTM with prior work on Youtube Celebrities Face dataset. Considering among the state-of-the-art work, only TT-RNN \cite{yang2017tensor}, including TT-GRU and TT-LSTM, reports both the accuracy and compression ratio on this dataset, the comparison in Table \ref{tab:directly_ytc} is mainly between HT-LSTM and TT-based solutions. From this table it is seen that HT-LSTM achieves 72,818$\times$ compression ratio over the original uncompressed LSTM with much higher accuracy. Compared with the existing high-compression model TT-GRU, HT-LSTM has 4.1$\times$ fewer parameters but offering 8.1\% accuracy increase. 

\begin{table}[h]
\vskip -2mm
\def\arraystretch{1.15}
\centering
\begin{tabular}{|c|c|c|c|}  
\hline
\textbf{Model}  & \textbf{CR} & \textbf{\# Param.} & \textbf{Accuracy} (\%) \\
\hline
\hline
LSTM &  1$\times$  & 59M  & 33.2       \\
\hline
TT-GRU  & 17,723$\times$  & 3,328  & 80.0      \\
\hline
TT-LSTM & 17,388$\times$  & 3,392  & 75.5      \\
\hline
HT-LSTM (Ours)  & \textbf{72,818$\times$}   & \textbf{810}  & \textbf{88.1}      \\
\hline
\end{tabular}
\caption{
Performance of different RNN compression work on Youtube celebrities face dataset using end-to-end direct training. CR stands for compression ratios. Results of TT-LSTM and TT-GRU are reported in \protect\cite{yang2017tensor}.}
\label{tab:directly_ytc}
\vspace{-3mm}
\end{table}

Besides, on the same Youtube dataset we also compare HT-LSTM with several other reported works without using tensor decomposition method. As shown in Table \ref{tab:directly_ytc2}, among those works the state-of-the-art model is \cite{ortiz2013face}, which has the highest reported accuracy (80.8\%). Compared with that model, HT-LSTM achieves 7.3\% higher test accuracy.

\begin{table}[h]
\def\arraystretch{1.15}
\centering
\begin{tabular}{|l|c|}  
\hline
 \multicolumn{1}{|c|}{\textbf{Model}}  &  \textbf{Accuracy} (\%) \\
\hline
\hline
\cite{kim2008face}  &  71.2 \\
\hline
\cite{harandi2013dictionary}    & 73.9    \\
\hline
\cite{ortiz2013face}   & 80.8  \\
\hline
\cite{lu2015multi}      & 78.5     \\
\hline
HT-LSTM (Ours)  &  \textbf{88.1} \\
\hline
\end{tabular}
\caption{Comparison between HT-LSTM using end-to-end direct training and other models without using tensor decomposition on Youtube celebrities face dataset.}
\label{tab:directly_ytc2}
\vspace{-4.6mm}
\end{table}

\subsection{Training with Pre-trained CNNs}
Another set of our experiments is based on training strategy using pre-trained CNNs. To be specific, the pre-trained CNN first extracts the useful and compact features from the large-size raw input data, and then sends those captured features to RNN, As indicated in \cite{donahue2015long}, using the front-end CNN can not only reduce the required input vector size of RNN, but can also significantly improve the overall performance of the entire CNN+RNN model.

\textbf{Hyperparameter Setting.} In this part of experiments dropout rate is set as 0.5. The L2 regularization with coefficient 0.0001 is used, and the entire HT-LSTM model is trained using ADAM optimizer with batch size 16.

\textbf{UCF11 Dataset.} Consistent with \cite{pan2019compressing}, Inception-V3 
is selected as the the front-end CNN feature extractor, whose output is a flattened size-2,048 feature vector. For our proposed HT-LSTM, this feature vector, as the input to the model, is reshaped to a tensor of size $8\times 8\times 8\times 4$. Similarly, the output vector is reshaped to a tensor of size $4\times 8\times 8\times 8$. In addition, the ranks for all the nodes are set as 4. 

Table \ref{tab:cnn_ucf11} summarizes the test accuracy of different models over UCF11 dataset. It is seen that with pre-trained CNN model as front-end feature extractor, HT-LSTM achieves 98.1\% accuracy, which is 3.5\% higher than the best reported accuracy from the state of the art. Compared with the TR-LSTM using the same front-end CNN, HT-LSTM achieves 4.3\% accuracy increase. Meanwhile, being compressed from the same vanilla LSTM, HT-LSTM model brings very high compression ratio as 12,945; while the compression ratio of TR-LSTM is only 25.

\begin{table}[h]
\vskip -2mm
\def\arraystretch{1.15}
\centering
\begin{tabular}{|l|c|}  
\hline
 \multicolumn{1}{|c|}{\textbf{Model}}  &  \textbf{Accuracy} (\%) \\
 \hline
 \hline
\cite{wang2015action}  &  84.2 \\
\hline
\cite{sharma2015action}    &  86.0   \\
\hline
\cite{cho2014robust}   &  88.0 \\
\hline
\cite{gammulle2017two}      &    94.6  \\
\hline

CNN + LSTM \cite{pan2019compressing} & 92.3 \\
\hline
CNN + TR-LSTM \cite{pan2019compressing}  &  93.8 \\
\hline
CNN + HT-LSTM (Ours)  &  \textbf{98.1} \\

\hline
\end{tabular}
\caption{Comparison between HT-LSTM using front-end pre-trained CNN and other related work on UCF11 dataset.}
\label{tab:cnn_ucf11}
\vspace{-4mm}
\end{table}

\textbf{HMDB51 Dataset.} HMDB51 dataset \cite{kuehne2011hmdb} contains 6,849 video clips that belong to 51 action categories, where each of them consists of more than 101 clips. Again, for the experiment on this dataset we use Inception-V3 as the pre-trained CNN model, whose extracted feature is flattened to a length-2,048 vector. Reshaped from this vector, the input tensor has size of $8\times 8\times 8\times 4$. We also reshape the output vector to a tensor of size $4\times 8\times 8\times 8$. Meanwhile, the ranks of all the nodes are set as 4.

Table \ref{tab:cnn_hmdb51} summarizes the performance of CNN-aided HT-LSTM and other related work on this dataset. It is seen that HT-LSTM achieves 64.2\% accuracy, which obtains 0.4\% increase than the recent TR-LSTM. Meanwhile, the compression ratio brought by HT-LSTM is 12,945, which is much higher than the 25$\times$ parameter reduction in TR-LSTM. 

It is worth noting that the state-of-the-art work \cite{carreira2017quo} achieves a higher accuracy (66.4\%) than HT-LSTM. This performance is based on using two streams of input (RGB images and optical flow); while HT-LSTM does not utilize optical flow information of the video. When only RGB information of the video is sent to the models, HT-LSTM can achieve very competitive classification performance as compared to the prior work.

\begin{table}[h]
\vskip -2mm
\def\arraystretch{1.15}
\centering
\begin{tabular}{|l|r|}  
\hline
 \multicolumn{1}{|c}{\textbf{Model}}  &   \multicolumn{1}{|c|}{\textbf{Accuracy} (\%)} \\
\hline
\hline
\cite{wang2015action}    &  63.2   \\
\hline
\cite{feichtenhofer2016convolutional} & 56.8 \\
\hline
\multirow{2}{*}{\cite{carreira2017quo}}   &  RGB + Flow: \textbf{66.4}\\
& RGB: 49.8\\
\hline
CNN + LSTM \cite{pan2019compressing} & 62.9 \\
\hline
CNN + TR-LSTM \cite{pan2019compressing}  &  63.8 \\
\hline
CNN + HT-LSTM (Ours)  &  64.2 \\
\hline
\end{tabular}
\caption{Comparison between HT-LSTM using front-end pre-trained CNN and other related work on HMDB51 dataset.}
\label{tab:cnn_hmdb51}
\vskip -5mm
\end{table}

\section{Conclusion}

In this paper, we propose a new RNN compression approach using Hierarchical Tucker (HT) decomposition. The HT-based RNN models exhibit strong hierarchical structure as well as low storage and computational costs. Our experiments on different datasets show that, our proposed HT-LSTM models significantly outperform the state-of-the-art compressed RNN models in terms of both compression ratio and test accuracy.

\newpage

\bibliographystyle{named}
\bibliography{ijcai20}

\end{document}